\documentclass[11pt,a4paper]{article}
\usepackage[hyperref]{naaclhlt2018}
\usepackage{times}
\usepackage{latexsym}
\usepackage{graphicx}
\usepackage{amsmath}
\usepackage{amsfonts}

\usepackage{url}

\aclfinalcopy 

\title{NLITrans at SemEval-2018 Task 12: Transfer of Semantic Knowledge for Argument Comprehension}

\author{Timothy Niven {\normalfont and} Hung-Yu Kao \\
  National Cheng Kung University \\
  Tainan, Taiwan \\
  {\tt tim.niven.public@gmail.com}, \ {\tt hykao@mail.ncku.edu.tw}}

\date{}

\begin{document}
\maketitle
\begin{abstract}
  The Argument Reasoning Comprehension Task requires significant language understanding 
    and complex reasoning over world knowledge.
  We focus on transfer of a sentence encoder to bootstrap more complicated models
    given the small size of the dataset.
  Our best model uses a pre-trained BiLSTM to encode input sentences, 
    learns task-specific features for the argument and warrants,
    then performs independent argument-warrant matching.
  This model achieves mean test set accuracy of 64.43\%.
  Encoder transfer yields a significant gain to our best model over random initialization.
  Independent warrant matching effectively doubles the size of the dataset 
    and provides additional regularization.
  We demonstrate that regularization comes from ignoring statistical correlations 
    between warrant features and position.
  We also report an experiment with our best model that only matches warrants to reasons, ignoring claims.
  Relatively low performance degradation suggests that 
    our model is not necessarily learning the intended task.
\end{abstract}

\section{Introduction}

The Argument Reasoning Comprehension Task (ARCT) ~\cite{habernal.et.al.2018.NAACL.arct} addresses a significant open problem in argumentation mining: 
  connecting reasons and claims via inferential licenses, called warrants ~\cite{Toulmin:1958}.
Warrants are a form of shared world knowledge and are mostly implicit in argumentation.
This makes it difficult for machine learning algorithms to discover arguments,
  as they must acquire and use this knowledge to identify argument components and their relations.
ARCT isolates the reasoning step by not requiring warrants to be discovered.
A correct warrant $\mathbf{W}$ and an incorrect alternative $\mathbf{A}$ are given, 
  and the correct one must be predicted given the corresponding claim $\mathbf{C}$ and reason $\mathbf{R}$.

\begin{figure}[t]
\centering
\includegraphics[width=0.4\textwidth]{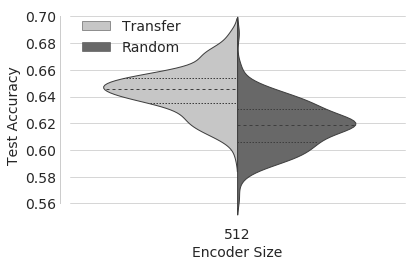}
\caption{Benefit of transfer to our best model, COMP.
Distributions come from 200 runs with different random seeds.
Mean accuracy for transfer (64.43\%) is higher than random (61.81\%)
  and is significant with $p = 9.68 \times 10^{-41}$.}
\end{figure}

However, this does not eliminate the need for other forms of world knowledge.
Consider the following example from the test set:

\vspace{8pt}

\begin{tabular}{p{0.03\textwidth}p{0.37\textwidth}}
  $\mathbf{C}$ & Google is not a harmful monopoly \\
  $\mathbf{R}$ & People can choose not to use Google \\
  $\mathbf{W}$ & Other search engines do not re-direct to Google \\
  $\mathbf{A}$ & All other search engines re-direct to Google \\
\end{tabular}

\vspace{8pt}

\noindent
It is required to know how consumer choice and web re-directs relate to the concept of monopoly in this context,
  and that Google is a search engine.

\begin{figure}[t]
\centering
           \includegraphics[width=0.3\textwidth]{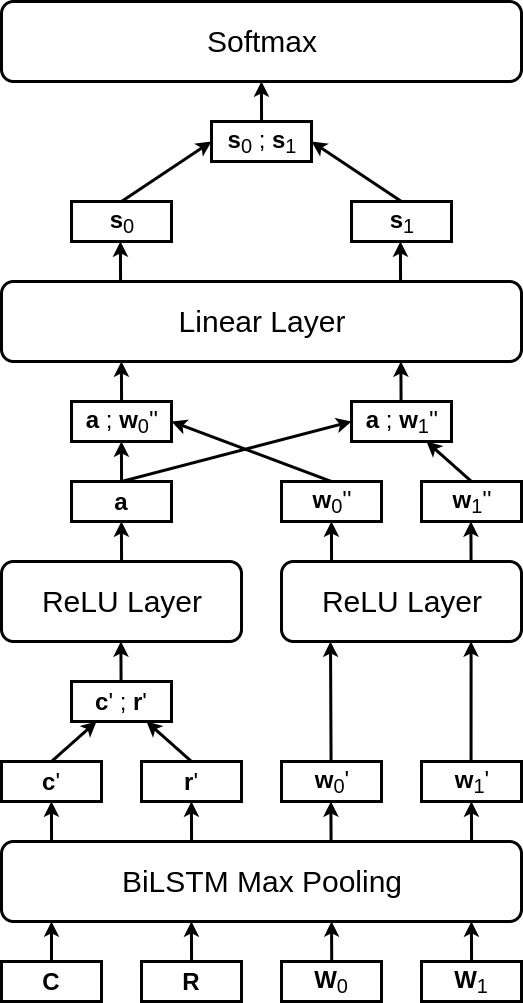}
\caption{
COMP model architecture.
}
\end{figure}

We do not attempt to address these other forms of world knowledge.
Given the small size of the dataset
  we focus on transfer of semantic knowledge in the form of a sentence encoder
  to bootstrap inference over learned features.
Following Conneau et al. ~\shortcite{DBLP:journals/corr/ConneauKSBB17}, 
  we pre-train a BiLSTM encoder with max pooling on natural language inference (NLI) data 
    ~\cite{DBLP:journals/corr/WilliamsNB17, DBLP:journals/corr/BowmanAPM15}. 
Their results indicate transfer from the NLI domain to be useful. 
They hypothesized that due to the challenging nature of the task,
  successful encoders must necessarily learn good semantic representations.
However, Nie et al. ~\shortcite{DBLP:journals/corr/abs-1710-04334} argue that
  due to the relatively easy nature of their out of domain generalization tasks 
  (sentiment classification and textual similarity),
  they did not sufficiently demonstrate that deep semantic understanding had been learned.

In this respect our work extends the results of Conneau et al. ~\shortcite{DBLP:journals/corr/ConneauKSBB17}.
They performed transfer by passing encoded sentence vectors to a logistic regression classifier.
Our implementation of this model demonstrated very poor performance on ARCT.
However, we experiment with a more complicated model (Figure 2)  
  which significantly benefits from transfer (Figure 1).
We therefore extend previous results to demonstrate the utility of this technique
  not only for a more semantically challenging task,
  but also a more complicated neural network architecture.

A key feature of our model is independent warrant classification
  which effectively doubles the size of the dataset.
We demonstrate that it also provides regularization due to 
  ignoring statistical correlations between warrant features and position.

Finally, we experiment with a version of our model 
  that only matches reasons to warrants, ignoring claims.
The relatively low drop in performance suggests that our model may not necessarily be learning the intended task.

\section{System Description}

\subsection{COMP Model}

A diagram of our best model which we call COMP is given in Figure 2.
The key idea is to learn independent features for argument components
  and then perform independent warrant matching.

The inputs are word vectors for the claim $\mathbf{C}$, reason $\mathbf{R}$, and warrants $\mathbf{W}_0$ and $\mathbf{W}_1$.
We use GloVe embeddings with 300 dimensions pre-trained on 640B tokens ~\cite{pennington2014glove}.
First, a bi-directional LSTM ~\cite{hochreiter1997long} with max pooling learns semantic representations of the input sentences.

\setlength{\belowdisplayskip}{0pt} \setlength{\belowdisplayshortskip}{0pt}
\setlength{\abovedisplayskip}{0pt} \setlength{\abovedisplayshortskip}{0pt}

\begin{align*}
\mathbf{c}' &= \text{BiLSTM}_{\max}(\mathbf{C})  \\
\mathbf{r}' &= \text{BiLSTM}_{\max}(\mathbf{R}) \\
\mathbf{w}_0' &= \text{BiLSTM}_{\max}(\mathbf{W}_0) \\
\mathbf{w}_1' &= \text{BiLSTM}_{\max}(\mathbf{W}_1) \\
\end{align*}

\noindent
Dropout ~\cite{JMLR:v15:srivastava14a} is then applied to these vectors.
If $d$ is the encoder size then each vector is of dimension $2d$
  due to the concatenation of forward and backward LSTMs.

Parameter matrix $\mathbf{U} \in \mathbb{R}^{4d \times h}$ with ReLU activation ~\cite{Nair2010RectifiedLU}
  learns argument specific feature vectors of length $h$ from the concatenation of the claim and reason.
Parameter matrix $\mathbf{V} \in \mathbb{R}^{2d \times h}$ with ReLU activation
  learns specific features for each warrant independently.
Biases are omitted for clarity.

\begin{align*}
\mathbf{a} &= \text{ReLU}(\mathbf{U} [\mathbf{c}' ; \mathbf{r}']) \\
\mathbf{w}_0'' &= \text{ReLU}(\mathbf{V}\mathbf{w}_0') \\
\mathbf{w}_1'' &= \text{ReLU}(\mathbf{V}\mathbf{w}_1') \\
\end{align*}

\noindent
Dropout is then applied to these vectors prior to classification.
Parameter vector $\mathbf{z} \in \mathbb{R}^{2h}$ is used to independently determine a matching score for each argument-warrant pair.
The scores are concatenated and passed through softmax to determine a probability distribution $\hat{\mathbf{y}}$ over the two warrants.
Cross entropy is then used to calculate loss with respect to the gold label $y$.

\begin{align*}
s_0 &= \mathbf{z}^\top [\mathbf{a} ; \mathbf{w}_0''] \\
s_1 &= \mathbf{z}^\top [\mathbf{a} ; \mathbf{w}_1''] \\
\hat{\mathbf{y}} &= \text{softmax}([s_0 ; s_1]) \\
J(\mathbf{\theta}) &= \text{CE}(\hat{\mathbf{y}}, y)
\end{align*}

\begin{table*}[t]
\centering
\begin{tabular}{lrrrr}
  \bf Encoder Size  & \bf Random & \bf Transfer & \bf Difference & \bf Significance (p) \\
  \hline
  2048 & 0.5975 & 0.5942 & -0.55 \% & $1.72 \times 10^{-1}$ \\
  1024 & 0.6058 & 0.6025 & -0.54 \% & $1.05 \times 10^{-1}$ \\
  512  & 0.6181 & \bf 0.6443 & \bf +4.24 \% & $9.68 \times 10^{-41}$ \\
  300  & 0.6285 & 0.6260 & -0.40 \% & $1.41 \times 10^{-1}$ \\
  100  & \bf 0.6310 & 0.6329 & +0.30 \% & $2.89 \times 10^{-1}$  \\
\end{tabular}
\caption{Transfer results for our COMP model with different encoder sizes.
Learning rates and dropout are tuned to specific encoder sizes.
All other hyperparameters are the same.
Results are the mean test set accuracy of 200 runs from different random seeds.
``Difference'' shows the percentage change of transfer relative to random.}
\end{table*}

\subsection{Training Details}

Pre-training BiLSTMs was done according to the specifications of Conneau et al. 
  ~\shortcite{DBLP:journals/corr/ConneauKSBB17}.
For all ARCT models we followed their annealing and early stopping strategy:
  after each epoch, if development set accuracy does not improve
  the learning rate is reduced by a factor of five.
Training stops when the learning rate drops below $1 \times 10^{-5}$.
This algorithm was found to outperform a steadily decaying learning rate.
Adam ~\cite{DBLP:journals/corr/KingmaB14} was used for optimization.

We used grid search to find our best parameters.
Best results were achieved with a batch size of 16.
Dropout with $p=0.1$ was found to be superior to $L2$-regularization.
For the COMP model, a hidden representation size of $512$ worked best.
Tuning word embeddings was found to overfit compared to freezing them.
However, tuning the transferred encoder was far superior to freezing for the COMP model.

We did not find reducing the learning rate on the encoder helped transfer.
Bowman et al. ~\shortcite{DBLP:journals/corr/BowmanAPM15} 
  also transfered AdaDelta accumulators along with an encoder
  on the principle that lowering the starting learning rate should help avoid blowing away transferred knowledge.
Our results rather agree with Mou et al. ~\shortcite{DBLP:journals/corr/MouMYLXZJ16} 
  who also found that learning rate reduction did not help transfer.

Our code is publicly available,
  including scripts to reproduce our results.\footnote{\href{https://github.com/IKMLab/arct}{https://github.com/IKMLab/arct}}

\subsection{Submission}

Our submission ``NLITrans'' was our COMP model with a transferred encoder of dimension 2048.
The principal learning rate was 0.002
  and we tuned embeddings at their own rate of 0.0002.
The encoder was tuned at the principal rate.
Hidden representation size was 512.

Our submission test set accuracy of 59.0\% achieved fourth place.
Following Riemers and Gurevych ~\shortcite{DBLP:journals/corr/ReimersG17a}, 
  we consider a single run an insufficient indication of the performance of a model
  due to the variation resulting from random initialization.
Evaluation over 20 runs with different random seeds
  revealed our entry was close to the mean for this configuration of 59.24\%.

\subsection{Best Configuration}

Extended post-competition tuning on the development set 
  revealed better hyperparameter settings
  that boost the generalization ability of our COMP model.
Specifically, we freeze word embeddings and use an encoder of size 512.
On the test set this configuration achieves a mean accuracy of 64.43\% over 200 random initializations.

\section{Analysis}

\subsection{Transfer Performance}

We measure the performance of transfer by comparison with random initialization.
The results in Table 1 show the performance of different encoder sizes for our best COMP model.
Learning rate and dropout are re-tuned via grid search for each encoder size.
More investigation is required, 
 however these results suggest that the success of transfer depends on 
 finding an optimal encoder size for a given model.
We note that the best random performance came from the smallest encoder size,
  confirming that transfer is helping us bootstrap the use of more complicated models.

\begin{table}[t]
\begin{center}
\begin{tabular}{llrrr}
\hline 
\bf Model & \bf Dataset & \bf Train & \bf Test & \bf Overfit \\ 
\hline
COMP & Full & 0.8807 & 0.6443 & 36.69 \% \\
     & Half & 0.8925 & 0.6332 & 40.95 \% \\
     & Unbal. & 0.9109 & 0.6353 & 43.38 \% \\
\hline
CORR & Full & 0.8155 & 0.5912 & 37.94 \% \\
     & Half & 0.8287 & 0.5649 & 46.70 \% \\
     & Unbal. & 0.9368 & 0.5750 & 62.92 \% \\
\hline
\end{tabular}
\end{center}
\caption{\label{font-table} Comparison of independent (COMP) and correlated (CORR) models
on full, half, and unbalanced (Unbal.) datasets.
COMP has 6,541,057 parameters and CORR has 6,543,106.}
\end{table}

\subsection{Independent Warrant Classification}

Ablation studies showed that independent warrant classification is a significant advantage.
For comparison, we built a model that considers both warrants and the argument together
  by replacing parameter vector $\mathbf{z}$ 
  with a matrix $\mathbf{Z} \in \mathbb{R}^{3h \times 2}$.
We call this model CORR, as it considers correlations between warrant features and position.
The scores for the warrants are then calculated as

\begin{align*}
  \mathbf{s} = \mathbf{Z} [\mathbf{a} ; \mathbf{w}_0'' ; \mathbf{w}_1'']
\end{align*}

\vspace{12pt}

\noindent
The results in Table 2 demonstrate poorer generalization and increased overfitting 
  relative to the independent model on the full dataset.

Independent warrant classification effectively doubles the size of the dataset.
It can be seen that two separate multiplications of argument-warrant vectors with $\mathbf{z}$
  lead to two backpropagated supervision signals with each data sample.
To quantify this effect we evaluated the independent model trained on
  a randomly sampled half of the the training set.
It still generalized better than CORR on twice the data (Table 2).

We hypothesized that additional regularization follows from 
  ignoring statistical correlations between warrant features and position.
To investigate this hypothesis, we picked an obvious linguistic phenomenon 
  and looked at the statistics of its occurrence at each warrant position.
We used SpaCy's dependency parser to identify tokens with the negation relation to its head.
Results showed negation is ubiquitous in this dataset, covering approximately 70\% of training samples -
  perhaps reflecting a natural way to generate pairs of conflicting warrants.
Whilst the warrant with negation is correct half of the time in the training set,
  negation in position one is slightly more likely to be correct than position zero (26\% to 25\%).

To quantify model susceptibility to such correlations 
  we created an unbalanced training set in which 
  all correct warrants with negation occur in position one.
This resulted in relabeling 300 warrants.
We randomly relabeled the same amount of warrants without negation to position zero 
  to re-balance the dataset.
The results (Table 2) confirm that the CORR model is more susceptible to this change,
  resulting in a large overfit,
  whilst the generalization ability of the independent model is essentially unaffected.

\begin{figure}[t]
\centering
\includegraphics[width=0.4\textwidth]{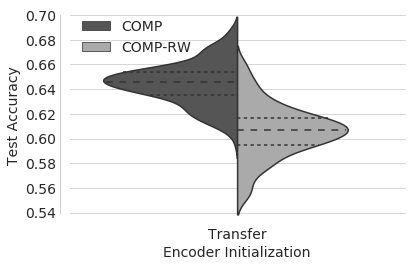}
\caption{Results of our COMP-RW model that doesn't consider the claim,
  compared to our best model COMP.
Distributions are calculated from 200 runs with different random seeds.
The mean for COMP is 64.43\%, compared to 60.60\% for COMP-RW.
}
\end{figure}

\subsection{Matching Warrants to Reasons}

In the following example a position toward the claim seems to be embedded in the reason.

\vspace{8pt}

\begin{tabular}{p{0.03\textwidth}p{0.37\textwidth}}
  $\mathbf{C}$ & Comment sections have not failed \\
  $\mathbf{R}$ & They add a lot to the piece and I look forward to reading comments \\
  $\mathbf{W}$ & Comments sections are a welcome distraction from my work \\
  $\mathbf{A}$ & Comments sections always distract me from my work \\
\end{tabular}

\vspace{8pt}

\noindent
Cases such as this may provide an alternative learning signal and lead our model to stray from the intended task.
For example it might be possible to correctly classify this example
  by comparing the sentiment of the warrants to that of the reason.

To quantify this effect we experimented with a model 
  that considers only the reasons and warrants, called COMP-RW.
Since we do not input the claim we resize $\mathbf{U}$ 
  from $\mathbb{R}^{4d \times h}$ to $\mathbb{R}^{2d \times h}$.
We use an encoder of size 640 to balance this reduction
  which evens the parameter count for a fair comparison.
Figure 3 shows the relative performance of COMP-RW versus our best COMP model.
Test set accuracy deteriorates from 64.43\% to 60.60\%.
This suggests there is enough signal coming from the reasons alone
  to achieve approximately two thirds of what our model is capable of above random guessing.
We therefore suspect our model is not necessarily learning the intended task.

\section{Conclusion}

Our entry NLITrans achieved test set accuracy of 59.0\% for fourth place,
  close to the mean for its configuration of 59.24\% over 20 random initializations.
Extended post-competition tuning on the development set 
  led us to a superior configuration of our COMP model 
  that achieved a mean of 64.43\%.
Transfer of an encoder pre-trained on NLI data resulted in a 4.24\% boost to test set accuracy for this model.
This extends previous results with this transfer technique,
  demonstrating its effectiveness 
  in a more complicated neural network architecture,
  and for a much more semantically challenging task.
An outstanding question is whether there is an optimal encoder size for transfer 
  given a specific architecture, 
  and how to efficiently and reliably find it.
Independent argument-warrant matching proved to be beneficial,
  doubling the effective size of the dataset and providing additional regularization.
We demonstrated that regularization comes from ignoring the correlations between warrant features and position.
Adapting our model to ignore the claims 
  resulted in a relatively low drop in performance,
  suggesting our model is not necessarily learning the intended task.
We leave a more thorough analysis of this phenomenon for future work.

\bibliography{naaclhlt2018}

\begin{thebibliography}{13}
\expandafter\ifx\csname natexlab\endcsname\relax\def\natexlab#1{#1}\fi

\bibitem[{Bowman et~al.(2015)Bowman, Angeli, Potts, and
  Manning}]{DBLP:journals/corr/BowmanAPM15}
Samuel~R. Bowman, Gabor Angeli, Christopher Potts, and Christopher~D. Manning.
  2015.
\newblock \href {http://arxiv.org/abs/1508.05326} {A large annotated corpus for
  learning natural language inference}.
\newblock \emph{CoRR}, abs/1508.05326.

\bibitem[{Conneau et~al.(2017)Conneau, Kiela, Schwenk, Barrault, and
  Bordes}]{DBLP:journals/corr/ConneauKSBB17}
Alexis Conneau, Douwe Kiela, Holger Schwenk, Lo{\"{\i}}c Barrault, and Antoine
  Bordes. 2017.
\newblock \href {http://arxiv.org/abs/1705.02364} {Supervised learning of
  universal sentence representations from natural language inference data}.
\newblock \emph{CoRR}, abs/1705.02364.

\bibitem[{Habernal et~al.(2018)Habernal, Wachsmuth, Gurevych, and
  Stein}]{habernal.et.al.2018.NAACL.arct}
Ivan Habernal, Henning Wachsmuth, Iryna Gurevych, and Benno Stein. 2018.
\newblock \href {https://arxiv.org/abs/1708.01425} {The argument reasoning
  comprehension task: Identification and reconstruction of implicit warrants}.
\newblock In \emph{Proceedings of the 2018 Conference of the North American
  Chapter of the Association for Computational Linguistics: Human Language
  Technologies}, page (to appear), New Orleans, LA, USA. Association for
  Computational Linguistics.

\bibitem[{Hochreiter and Schmidhuber(1997)}]{hochreiter1997long}
Sepp Hochreiter and J{\"u}rgen Schmidhuber. 1997.
\newblock Long short-term memory.
\newblock \emph{Neural computation}, 9(8):1735--1780.

\bibitem[{Kingma and Ba(2014)}]{DBLP:journals/corr/KingmaB14}
Diederik~P. Kingma and Jimmy Ba. 2014.
\newblock \href {http://arxiv.org/abs/1412.6980} {Adam: {A} method for
  stochastic optimization}.
\newblock \emph{CoRR}, abs/1412.6980.

\bibitem[{Mou et~al.(2016)Mou, Meng, Yan, Li, Xu, Zhang, and
  Jin}]{DBLP:journals/corr/MouMYLXZJ16}
Lili Mou, Zhao Meng, Rui Yan, Ge~Li, Yan Xu, Lu~Zhang, and Zhi Jin. 2016.
\newblock \href {http://arxiv.org/abs/1603.06111} {How transferable are neural
  networks in {NLP} applications?}
\newblock \emph{CoRR}, abs/1603.06111.

\bibitem[{Nair and Hinton(2010)}]{Nair2010RectifiedLU}
Vinod Nair and Geoffrey~E. Hinton. 2010.
\newblock Rectified linear units improve restricted boltzmann machines.
\newblock In \emph{ICML}.

\bibitem[{Nie et~al.(2017)Nie, Bennett, and
  Goodman}]{DBLP:journals/corr/abs-1710-04334}
Allen Nie, Erin~D. Bennett, and Noah~D. Goodman. 2017.
\newblock \href {http://arxiv.org/abs/1710.04334} {Dissent: Sentence
  representation learning from explicit discourse relations}.
\newblock \emph{CoRR}, abs/1710.04334.

\bibitem[{Pennington et~al.(2014)Pennington, Socher, and
  Manning}]{pennington2014glove}
Jeffrey Pennington, Richard Socher, and Christopher~D. Manning. 2014.
\newblock \href {http://www.aclweb.org/anthology/D14-1162} {Glove: Global
  vectors for word representation}.
\newblock In \emph{Empirical Methods in Natural Language Processing (EMNLP)},
  pages 1532--1543.

\bibitem[{Reimers and Gurevych(2017)}]{DBLP:journals/corr/ReimersG17a}
Nils Reimers and Iryna Gurevych. 2017.
\newblock \href {http://arxiv.org/abs/1707.09861} {Reporting score
  distributions makes a difference: Performance study of lstm-networks for
  sequence tagging}.
\newblock \emph{CoRR}, abs/1707.09861.

\bibitem[{Srivastava et~al.(2014)Srivastava, Hinton, Krizhevsky, Sutskever, and
  Salakhutdinov}]{JMLR:v15:srivastava14a}
Nitish Srivastava, Geoffrey Hinton, Alex Krizhevsky, Ilya Sutskever, and Ruslan
  Salakhutdinov. 2014.
\newblock \href {http://jmlr.org/papers/v15/srivastava14a.html} {Dropout: A
  simple way to prevent neural networks from overfitting}.
\newblock \emph{Journal of Machine Learning Research}, 15:1929--1958.

\bibitem[{Toulmin(1958)}]{Toulmin:1958}
Stephen~E. Toulmin. 1958.
\newblock \emph{The Uses of Argument}.
\newblock Cambridge University Press.

\bibitem[{Williams et~al.(2017)Williams, Nangia, and
  Bowman}]{DBLP:journals/corr/WilliamsNB17}
Adina Williams, Nikita Nangia, and Samuel~R. Bowman. 2017.
\newblock \href {http://arxiv.org/abs/1704.05426} {A broad-coverage challenge
  corpus for sentence understanding through inference}.
\newblock \emph{CoRR}, abs/1704.05426.

\end{thebibliography}
\bibliographystyle{acl_natbib}

\end{document}